\documentclass[letterpaper, 10 pt, conference]{ieeeconf}
\IEEEoverridecommandlockouts                              
\newcommand{\orcid}[1]{\href{https://orcid.org/#1}{\textcolor[HTML]{A6CE39}{\aiOrcid}}}

\usepackage{subfiles}

\usepackage{cite}
\usepackage[caption=false,font=normalsize,labelfont=sf,textfont=sf]{subfig}

\usepackage{hyperref}
\usepackage{academicons}
\usepackage{xcolor}

\usepackage{amsmath}
\usepackage{amsfonts}

\usepackage{graphicx}
\graphicspath{{media/}}
\usepackage{pgfplots}
\usepackage{algorithm2e}
\pgfplotsset{compat=1.18}

\definecolor{mycolor1}{HTML}{1f77b4}
\definecolor{mycolor2}{HTML}{ff7f0e}
\definecolor{mycolor3}{HTML}{2ca02c}
\definecolor{mycolor4}{HTML}{d62728}
\definecolor{mycolor5}{HTML}{9467bd}
\definecolor{mycolor6}{HTML}{8c564b}
\def\PlotThickness{2.5pt}
\def\LegendLineThickness{1.5pt}

\newcommand\LNorm[1]{\left\lVert#1\right\rVert}

\begin{document}

\title{\LARGE \bf Shaping Wind-Tunnel Airflow for Unmanned Aerial Vehicles using Online Learning}

\author{Ghadeer Elmkaiel and Michael Muehlebach
\thanks{G. Elmkaiel and M. Muehlebach are with the Learning and Dynamical Systems group at Max Planck Institute for Intelligent Systems, T\"ubingen 72076, Germany (e-mail: ghadeer.elmkaiel@tuebingen.mpg.de).}
\thanks{Accepted for publication at the IEEE/RSJ International Conference on Intelligent Robots and Systems. © 2026 IEEE. Personal use of this material is permitted.}
}
    
\maketitle

\begin{abstract}
   
    The development and testing of advanced aerial robots require experiments in controlled environments with tailored airflow profiles. This paper presents an online learning algorithm for controlling the complex airflow field in a multi-fan vertical wind tunnel. Our method combines a simplified physical model with iterative, measurement-based learning, enabling sample-efficient convergence to desired airflow distributions. We demonstrate the method's versatility by generating complex airflow, such as uniform, Gaussian, and parabolic profiles. Crucially, we show that our algorithm can produce an airflow profile specifically designed for passive soaring, greatly enhancing flight performance of a soaring robot. Variability, practical utility, and robustness of our approach are further highlighted by successful operation with a varying number of fans.
\end{abstract}

\section{Introduction}

\begin{figure}
    \centering
    \includegraphics[width=3.3in]{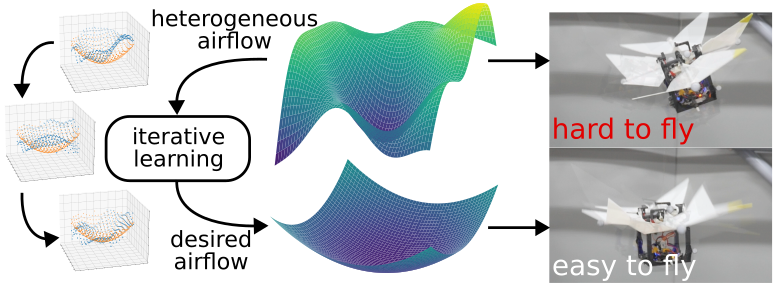}
    \caption{Our algorithm is capable of iteratively adapting the wind tunnel's airflow for robot soaring.}
    \label{fig:summary}
\end{figure}

A major promise of modern machine learning lies in enabling robots to interact with complex dynamical systems. Fluid dynamical systems are a prime example, exhibiting nonlinearities, critical transitions, and even chaos. A particularly compelling challenge in robotics is therefore the development of energy-efficient aerial robots that can harness, rather than fight, these complex fluid dynamics. 
However, designing, controlling, and testing underactuated soaring robots (see Fig.~\ref{fig:summary} and Fig.~\ref{fig:setup}) presents a significant challenge, and requires precise control of the surrounding airflow.
To facilitate research on such robots, we have developed a custom vertical wind tunnel to serve as a versatile experimental testbed. The challenge, however, lies in controlling this environment. The tunnel is actuated by seven propellers, which produce a turbulent and asymmetric flow field even under identical fan velocities (see Fig.~\ref{fig:initial_measurements}~(b)). This inherent asymmetry creates difficulties for conducting repeatable, high-fidelity experiments for aerodynamically sensitive robots. Therefore, a method to precisely and reliably shape the airflow field into a desired, stable profile is not just an improvement but a prerequisite for using this testbed for advanced robotics research.
In contrast to approaches that rely on computationally expensive fluid dynamics simulations, our work tackles this problem using online learning. We present an iterative algorithm to adapt the airflow relying only on measurement data, which has three key advantages: i) It directly relies on real-world force measurements from the wind tunnel, capturing all complex aerodynamic effects without an explicit model. ii) It incorporates a simplified, low-complexity physical model of the airflow to guide the learning process, making the process highly sample efficient. iii) It avoids the need to model turbulence, boundary conditions, and detailed propeller interactions. This paper demonstrates that our online learning scheme can rapidly converge to a variety of prespecified airflow distributions, and critically, can generate a specific parabolic profile that promotes stable flight for our soaring robot (see Fig.~\ref{fig:summary}).

\section{Related work}
The use of controlled airflow fields is a cornerstone of aerodynamic research and is becoming increasingly vital for the experimental testing of aerial robots \cite{russell2016wind, jindeog2003wind, bannwarth2018development}. Commercial systems, such as the multi-fan wind tunnels \cite{WindShape}, underscore the growing demand for generating specific weather conditions for testing Unmanned Aerial Vehicles (UAVs) \cite{flying_robot_1, flying_bird}. Academic research has also focused on developing these platforms, from large-scale open-jet facilities for aeroacoustic testing \cite{Acoustic_1, Acoustic_2} to "active grids" of fans designed to generate specific turbulent flows \cite{WindShape, wind_tunnel_2, wind_tunnel_3}. While these systems provide valuable platforms, the underlying challenge of actively controlling the shape of a turbulent airflow field into a precise, stable profile remains a complex problem at the intersection of fluid dynamics and control theory.
Historically, modeling such fluid systems has been the domain of high-fidelity, first-principle methods. These are typically based on computational fluid dynamics that solve the Navier-Stokes equations \cite{wang1980atmospheric, kundu2015fluid}. While accurate for analyzing phenomena like propeller-wing interactions \cite{fluid_1, fluid_2, fluid_3} or the complete flow field around a UAV \cite{fluid_4}, these methods are computationally prohibitive for physical testbeds. Their long computation times and the need for detailed mesh and boundary condition modeling make them fundamentally offline tools, unsuited for the rapid, on-the-fly adjustments needed in an experimental robotics workflow.

More recently, the robotics and machine learning communities have made significant advances in data-driven modeling of complex physical systems with physics-informed machine learning. A prominent example is the use of Physics-Informed Neural Networks (PINNs), which excel at learning solutions to partial differential equations directly from data \cite{raissi2017physics, willard2020integrating, raissi2019physics}. PINNs have been successfully applied to challenging fluid dynamics problems \cite{cai2021physics}, such as modeling vortex-induced vibrations and learning Lyapunov-stable flow predictions \cite{raissi2019deep, erichson2019physics}. In the control domain, PINNs have been used to learn a system's state-space model \cite{arnold2021state, ding2022pressing}. However, the primary limitation of these methods in our context is their reliance on extensive offline training with large datasets. Furthermore, they are not easily adaptable; if the physical system changes—due to hardware modifications or wear—the network must be retrained, making it inflexible for an evolving experimental setup.
Our work charts a different course, aligning with online, model-based iterative learning methods that have proven effective for controlling complex robotic systems with high sample efficiency. This paradigm, which combines a simplified physical model with direct real-world measurements \cite{ma2024stochastic}, has been successfully applied to challenging tasks such as learning to control agile quadcopter maneuvers\cite{lupashin2010simple, lambert2020nonholonomic}, controlling soft pneumatic robot arms \cite{ma2022learning, tobuschat2023data, zughaibi2021fast}, and in learning-based model predictive control for trajectory tracking \cite{sferrazza2020learning}. Instead of learning a complete, high-fidelity model offline, our algorithm uses a coarse physical model to provide a gradient that guides the learning process. 

We show the effectiveness of our algorithm on the example of controlling the airflow in a wind tunnel and generating airflow profiles suitable for flying with a soaring robot \cite{Soaring_robot_2}.
Our method complements a data-driven approach with physical prior knowledge, eliminating the need for complex simulations. The online learning combines measurement data with a coarse physical model that guides the learning and reduces the sample complexity.


\section{Experimental setup and problem formulation}
\label{sec:setup}
This section describes the experimental setup used in this work and introduces the problem of controlling the flow field in a wind tunnel. 
\subsection{Experimental setup}
\begin{figure*}[!t]
    \vspace{6pt} 
    \centering
    \includegraphics[width=6.4
    in]{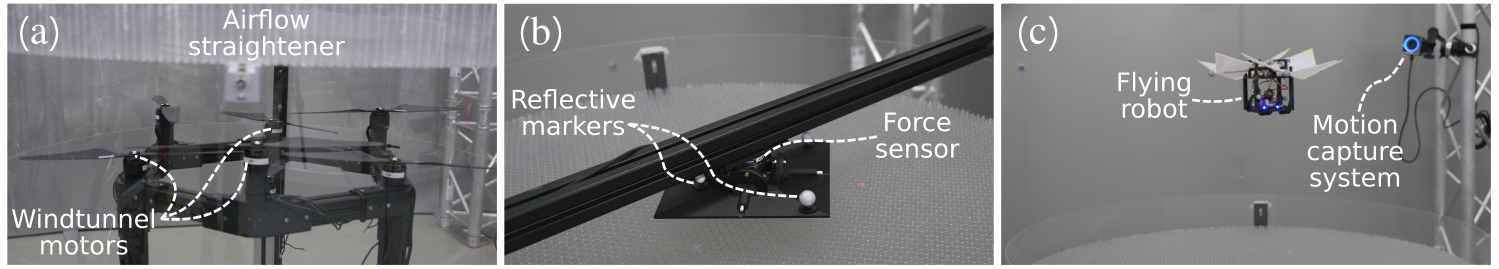}

    \caption{The figure summarizes the experimental setup used to test the algorithm. Panel (a) shows the core of the wind tunnel with the motors and honeycomb airflow straightener, panel (b) shows the ATI mini40 sensor with the plate that is used for the force/airflow measurements, and panel (c) shows the full setup with the flying robot, the wind tunnel, and the motion capture system.}
    \label{fig:setup}
\end{figure*}

\textbf{Wind tunnel:} 
Our setup consists of a custom-made vertical wind tunnel, which is actuated by seven brushless motors and propellers (can be controlled individually). Six propellers are positioned on the corners of a regular hexagon; three rotate clock-wise; the others rotate counter clock-wise. The last motor is positioned at the center. The motors are located in different horizontal planes, which allows for an overlap between the clock-wise and counter clock-wise propellers to reduce the possibility of creating large swirls (see Fig.~\ref{fig:setup}~(a)). Our goal is to control the vertical airflow generated by the propellers. The tunnel is equipped with a 14 mm honeycomb airflow straightener layer positioned directly above the propellers (visible in Fig.~\ref{fig:setup}~(a)). This straightener substantially eliminates the horizontal velocity components of the airflow, physically justifying the assumption that the aerodynamic force is perpendicular to the sensor plate.

\textbf{Airflow measurement setup:}
We measure the airflow indirectly by capturing the vertical force applied to a horizontal plate (see Fig.~\ref{fig:setup} (b)). The measurement of the reaction force has the advantage of providing an average over the surface area, effectively reducing noise due to turbulence. Furthermore, the force average over a small area is the value of interest for our application: testing UAVs and developing soaring robots (see Fig.~\ref{fig:setup}~(c)). It is important to highlight that our algorithm is easily adapted to other airflow measurement techniques, such as airspeed and pressure sensors. Moreover, for uniform airflow, the speed and the reaction force on the plate are directly related:
\begin{equation}
F_{\mathrm{plate}}(v_{\mathrm{air}})=\frac{1}{2}\rho A C_\mathrm{D} v_{\mathrm{air}}^2,
\label{eq:drag_force}
\end{equation}
where $\rho = 1.225\ \mathrm{kg/m^3}$ is the air density, $A=0.04\ \mathrm{m^2}$ the surface area of the plate perpendicular to the airflow, $C_\mathrm{D} = 1.15$ the drag coefficient, and $v_\mathrm{air}$ the airflow speed. 

To collect the airflow measurements, we connect the sensor to an aluminum bar and move the bar horizontally over the wind tunnel at a height of $150\ \mathrm{cm}$ above the propellers. We simultaneously track the position of the sensor with a motion capture system, resulting in an accurate measurement of the flow field above the rotating propellers (see Fig.~\ref{fig:setup}).

\subsection{Problem formulation and notation}
\begin{figure*}[!t]
    \centering
    \includegraphics[width=6in]{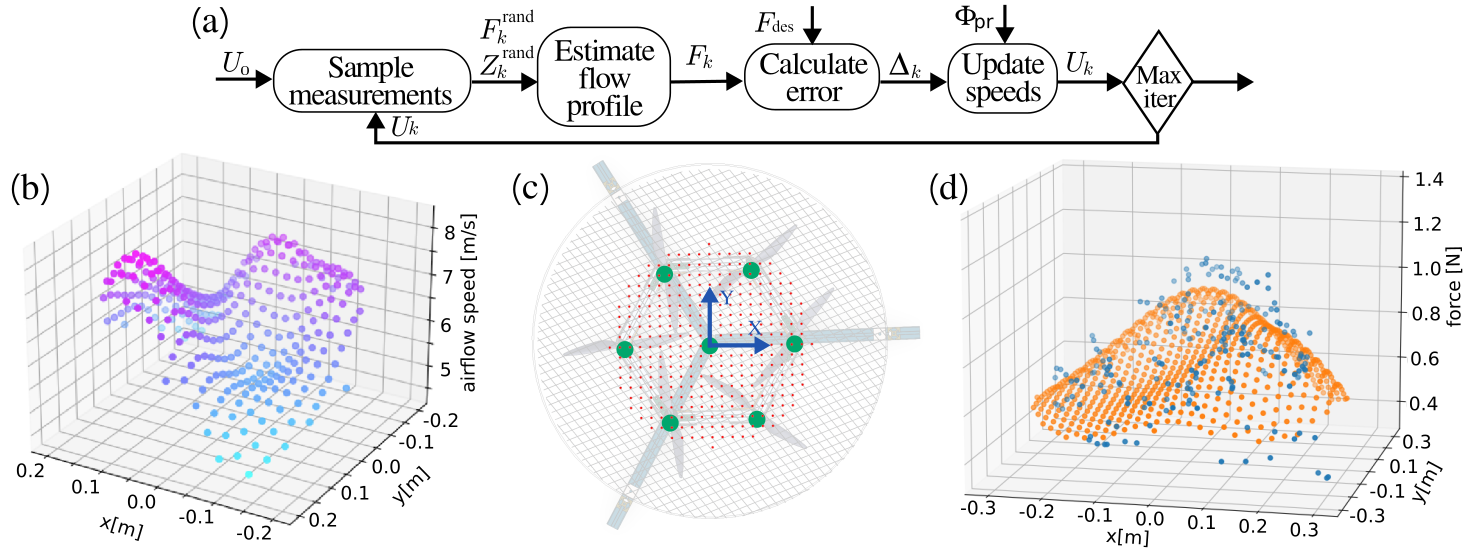}

    \caption{Panel (a) shows a flowchart of the algorithm. Panel (b) shows the initial airflow speeds over the wind tunnel when using the same rotational speed for all motors. Panel (c) highlights the position of the key points used in the online learning algorithm. The small red points are the key points, while the motors are highlighted with seven large green circles. There are 317 key points. Panel (d) displays an example of the non-parametric regression step (the blue points are the measurements, while the orange points correspond to the evaluation of the regression model at the key points).}
    \label{fig:initial_measurements}
\end{figure*}

The primary application for our wind tunnel is the experimental testing of UAVs and passive soaring robots.
In theory, running all motors at an identical speed should produce a symmetric airflow. In practice, minor physical imperfections in the motors, propellers, and overall setup result in a highly asymmetric flow field, as shown in Fig.~\ref{fig:initial_measurements} (b).
In the following, we will present an online learning algorithm that systematically compensates for these imperfections and controls the airflow to match predefined values. To that extent, we define a set of $N=317$ key points at which we evaluate the performance of the system, $\Omega=\ \{(\hat{x}_1,\hat{y}_1),\ (\hat{x}_2,\hat{y}_2),\ \dots,\ (\hat{x}_N,\hat{y}_N)\}$ where $\hat{z}_i = (\hat{x}_i, \hat{y}_i)$ specify the $xy$-coordinate of the key point $i$ in the wind tunnel. The key points used in the experiment are shown in Fig.~\ref{fig:initial_measurements} (c) and are all located at a height of $150\  \mathrm{cm}$ over the propellers. In an ideal case, direct force measurements at the key points would be desirable. However, this is exceedingly challenging in our experimental setting. To estimate the force measurements at the key points, we employ a non-parametric regression model, which enables us to extrapolate and reconstruct the data. This significantly reduces the experimental effort and will be further discussed in Section \ref{sec:methods}.

We denote a measurement by adding a \textit{bar} symbol, that is, $\bar Z = ((\bar{x}_1,\ \bar{y}_1),\ (\bar{x}_2,\ \bar{y}_2),\dots,\ (\bar{x}_N,\ \bar{y}_N))$ are the spatial coordinates of the selected measurements used in our algorithm, and $\bar F = (\bar{F}_1,\dots ,\bar{F}_N)$ are the corresponding force readings at those points. 

Our problem is formulated as follows: Find the motor speeds $U \in [0,1]\times...\times[0,1]$ that minimize the error $\mathrm{Err}(U)$ over the set of key points $\hat{z}\in \Omega$ (see Fig.~\ref{fig:initial_measurements} (c))
\begin{equation}
    \displaystyle\min_{U \in [0,1]^7} \mathrm{Err}(U) =\min_{U \in [0,1]^7} \frac{1}{N} \displaystyle\sum_{\hat{z}\in \Omega} (F(U,\hat{z})-F_\mathrm{des}(\hat{z}))^2,
    \label{eq:error_equation}
\end{equation}
where $F_\mathrm{des}(\hat{z})$ and $F(U,\hat{z})$ are the desired force value and the actual force value evaluated at the key point $\hat{z}$, respectively.

It is important to note that we minimize the force difference; however, as shown in \eqref{eq:drag_force}, the force is directly related to the airflow.


\section{Methodological overview: Algorithm, prior knowledge, and nonparametric regression}
\label{sec:methods}
In this section, we discuss the details of our algorithm. Furthermore, we describe how prior physical knowledge is incorporated in the learning process.
\subsection{Online learning algorithm}
For controlling the airflow and matching the prespecified distribution $F_\mathrm{des}(\hat{z})$, we apply the online learning algorithm given in Algorithm \ref{alg:online_learning_algorithm} (see Fig.~\ref{fig:initial_measurements} (a)). The algorithm starts by initializing the motors with the speeds $U_0$ (in our experiments, we typically initiate all motors with the same speed, which is 50\% of the maximum speed, that is, $U_0 = [0.5, ... , 0.5]$). The algorithm then continues as follows: i) Each iteration $k$ of the algorithm starts with moving the force sensor over the area of the wind tunnel, which results in the collection of $M$ measurements $F^\mathrm{rand}_k$over a random trajectory $Z^\mathrm{rand}_k$; ii) From those points, we extract a subset of $N$ measurements which are located closest to the key points $\bar{F}_{k,i}$, $\bar{z}_{k,i}$, where $\bar {z}_{k,i} \in \mathbb{R}^2, i= 1,\dots,N$. This approach reduces the randomness caused by the random choice of the trajectory and ensures that the subsampled points follow a uniform distribution over the operational area; iii) We use these measurements to perform a non-parametric Gaussian Process (GP) regression, resulting in an estimate of the force field over the entire surface $F_k(U_k,\cdot): \mathbb{R}^2 \to \mathbb{R}$; iv) We then calculate the error $\Delta_k \in \mathbb{R}^N$ between the estimated force and the desired force at the key points $\hat{z} \in \Omega$; v) Finally, we rely on a low-complexity physical model $\Phi_\mathrm{pr}(U,z)$, which will be detailed in the next paragraph, to update our motor speeds. The algorithm is repeated until convergence (see Algorithm \ref{alg:online_learning_algorithm}). We note that $ ^+$ refers to the Moore–Penrose pseudo inverse and $\lambda_k$ to the learning rate for iteration $k$. The algorithm can be viewed through the lens of a quasi-Newton method with an inexact model-based gradient.

\RestyleAlgo{ruled} 

\begin{algorithm}
\caption{Online Learning Algorithm}\label{alg:online_learning_algorithm}
\KwData{$U_0,\ N_\mathrm{Iter},\ \Omega, F_\mathrm{des},\ \lambda(\mathrm{step\ size}),\ {\Phi_\mathrm{pr}}(\mathrm{model})$}
\KwResult{$U_{N_\mathrm{Iter}+1}$}

\For{$k = 0\ \textbf{to} \ N_\mathrm{Iter}-1$}{
  \textbf{SET\_MOTORS\_SPEED}$(U_k)$;\\
    $F^\mathrm{rand}_k, Z^\mathrm{rand}_k \gets \textbf{MEASUREMENTS}$;\\
    $\bar{F}_k, \bar{Z}_k \gets \textbf{GET\_CLOSEST}(F^\mathrm{rand}_k, Z^\mathrm{rand}_k, \Omega)$;\\
  $F_k(U_k,\cdot) \gets GP(\bar{F}_k, \bar{Z}_k)$;\\
  ${\Delta_k} \gets \{(F_{\mathrm{des}}(\hat{z}_i)-F_{k}(U_k, \hat{z}_i))\}_{\hat{z}_i \in \Omega}$;\\
  $U_{k+1} \gets U_{k} +\lambda_{k} \dfrac {\partial {\Phi_\mathrm{pr}}}{\partial U}
\biggm\vert_{U_{k},\ \hat{z}}^{+}{\Delta_k}$;\\
}
\end{algorithm}

\subsection{Incorporating prior physical knowledge}

We first derive a simplified physical model for the airflow generated by a single motor, and then apply superposition to obtain an airflow model that accounts for all motors.
The airflow generated by the edge motors $j\in\{1,...,6\}$ at the point $z$, which has a distance $d_j(z)$ to the motor $j$, is described by the function:
\begin{equation}
\phi_{\mathrm{pr}(j)}(U_j, z) = P_\mathrm{h/l}(U_j)\exp\left(-d^2_j(z)/\sigma^2\right),
\label{eq:prior}
\end{equation}
where $P_\mathrm{h/l}(U_j)$ is a polynomial scaling function of degree two that maps the pulse-width modulated (PWM) command of motor $j$ to the force measured exactly on top of the motor. The polynomial for the higher motors is denoted by $P_\mathrm{h}(U_j)$, whereas $P_\mathrm{l}(U_j)$ is the polynomial for the lower motors. The parameter $\sigma$ is related to the size and shape of the force measurement surface, and the values of $P_\mathrm{h/l}(U_j)$ and $\sigma$ are found experimentally by fitting to a small batch of measurements (see Appendix \ref{sec:poly_fit}).
As for the central motor, the airflow generated at point $z$, which has a distance $d_c(z)$ to the center, is described by the function:
\begin{equation}
\phi_{\mathrm{pr(7)}}(U_7, z) = P_\mathrm{l}(U_7)(0.1+d^2_c(z))\exp\left(-d^2_c(z)/\sigma^2\right).
\label{eq:prior_central}
\end{equation}
Unexpectedly, increasing the central motor's speed primarily increases the airflow over the edge motors, as highlighted by \eqref{eq:prior_central}. This is likely due to the lower mounting position of the central motor and the complex aerodynamic interactions. This further illustrates the challenge in controlling the airflow.
We combine the individual models $\phi_{\mathrm{pr}(j)}(U_j, z)$ and obtain
\begin{equation}
 \Phi_{\mathrm{pr}}(U, z)=\displaystyle\sum_{j=1}^{N_{\mathrm{motors}}} \phi_{\mathrm{pr}(j)}(U_j, z),
\label{eq:prior_superposition}
\end{equation}
with $N_\mathrm{motors} = 7 $, which is used in Algorithm \ref{alg:online_learning_algorithm} to update the motor speeds.

It is clear that \eqref{eq:prior}, \eqref{eq:prior_central} and \eqref{eq:prior_superposition} are very coarse approximations of the underlying fluid dynamics. Nonetheless, we will show in the following that even a coarse model is enough to successfully guide the learning. This represents a key insight at the heart of our work.

\subsection{Nonparametric regression}
This section explains the nonparametric regression step in Algorithm \ref{alg:online_learning_algorithm}, which is used to extrapolate the force measurements to the key points. Due to the fact that the force typically follows a nonlinear distribution in space, we employ a GP regression model with a squared exponential kernel \cite{williams1995gaussian} (see also Fig.~\ref{fig:initial_measurements} (d)). Thus, we obtain the following function $F(U,\cdot): \mathbb{R}^2\to\mathbb{R}$ that describes the force field,
\begin{equation}
    F(U, z) =  {K}^\top(z, \bar{Z})\mathbf K(\bar{Z}, \bar{Z})^{-1}\bar F(U, \bar{Z}),
    \label{eq:nonparametric_regression}
\end{equation}
where $\bar{F}(U, \bar{Z})= (\bar{F}_1,\bar{F}_2,\dots,\bar{F}_N)$ denotes the subset of the force measurements at positions $\bar{Z} = (\bar{z}_1,\bar{z}_2,\dots,\bar{z}_N)$ when applying the PWM commands $U$, $\mathbf K(\bar{Z}, \bar{Z}) \in \mathbb{R}^{N\times N}$ is the matrix that contains the elements $k(\bar{z}_i,\bar{z}_j), i,j=1,\dots,N$, $K(z, \bar{Z}) \in \mathbb{R}^{N}$ is the vector that contains the elements $k(z, \bar{z}_i), i=1,\dots,N$, and where $k((x_1, y_1), (x_2, y_2)): \mathbb{R}^2\times\mathbb{R}^2\to\mathbb{R}$ denotes the kernel function.
We tested the performance of three different kernel functions (Gaussian, Laplace, and Tricube) on the task of extrapolating the airflow measurements and our results revealed that the Gaussian kernel had the best performance\footnote{To determine the optimal kernel function, we performed an evaluation using three distinct measurement datasets. Each dataset was collected by setting the motors to a different, representative speed configuration and recording 1000 measurement points over the wind tunnel. For each of the three datasets, we employed a 5-fold cross-validation scheme. In each fold, 200 points were used as the training set to build the regression model, and the remaining 800 points were used as the test set to evaluate its prediction performance.}.
The kernel function is therefore described as follows:
\begin{multline}\label{eq:kernel_function}
    k((x_1, y_1), (x_2, y_2)) =\\
    {}\exp{\left(- \dfrac{(x_1-x_2)^2+(y_1-y_2)^2}{2l^2} \right)},
\end{multline}
where $l$ denotes the length-scale. In our case, we found that the best length-scale parameter for the Gaussian kernel is $10\ \mathrm{cm}$, which is half the side length of our force measurement plate.


\section{Experimental results}
\label{sec:results}
\begin{figure*}[!t]
    \captionsetup[subfloat]{farskip=2pt,captionskip=2pt}
    \centering
    \includegraphics[width=6.5in]{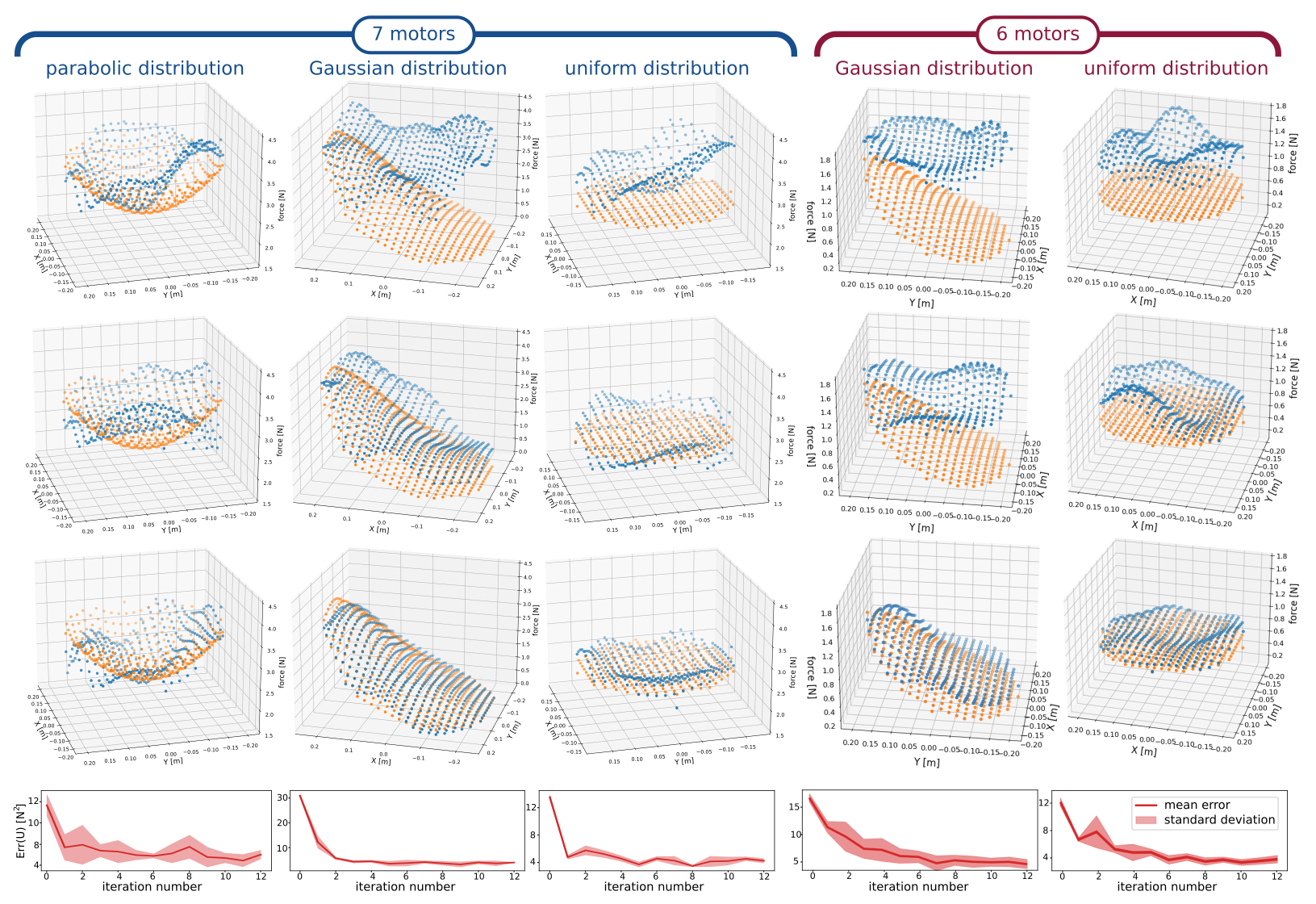}

    \caption{The figure shows the algorithm's performance for different desired airflow distributions. The first three columns represent the performance when using all the motors in the wind tunnel. The last two columns show the performance over two distributions when the central motor is disabled. The orange points show the target distribution, while the blue points show the measured distribution. Each row shows the evolution of the flow distribution (iteration zero, iteration one, avg over last five iterations). The slight differences among the images in the first row are due to variations in the collected measurement points and the resulting regression fits. The last row shows how the error decays for each airflow distribution. The shaded regions indicate the standard deviation over five different experiments, whereas the solid line denotes the mean.}
    \label{fig:algorithm_performance_all}
\end{figure*}

In this section, we show experimental results that demonstrate the performance of our algorithm in different situations. We performed multiple tests with different pre-specified force fields and found that in all cases, our algorithm was able to rapidly decrease the error defined in \eqref{eq:error_equation}, and match the pre-specified force distribution (see Fig.~\ref{fig:algorithm_performance_all}).

\textbf{Experiments' details:}
We ran our online learning algorithm (Algorithm \ref{alg:online_learning_algorithm}) as follows:
For each desired force distribution we performed five independent runs. In each run, we started by setting all motors to the same speed, which is $50\%$ of their maximum speed. We collected measurements by moving the sensor in a horizontal trajectory over the wind tunnel and recorded 1000 measurements and subsampled the $N=317$ points closest to the key points for the regression. In each experiment, we performed twelve iterations using a decaying learning rate $\lambda_k = 0.8/(k+1)$ where $k = 0,\dots,11$ is the iteration number.
We evaluate the algorithm on three target force distributions.

\begin{figure*}[!t]
    \centering
    \subfloat[]{
\begin{tikzpicture}[scale = 0.5]


\definecolor{crimson2143940}{RGB}{214,39,40}
\definecolor{darkgrey176}{RGB}{176,176,176}
\definecolor{darkorange25512714}{RGB}{255,127,14}
\definecolor{forestgreen4416044}{RGB}{44,160,44}
\definecolor{lightgrey204}{RGB}{204,204,204}
\definecolor{mediumpurple148103189}{RGB}{148,103,189}
\definecolor{orchid227119194}{RGB}{227,119,194}
\definecolor{sienna1408675}{RGB}{140,86,75}
\definecolor{steelblue31119180}{RGB}{31,119,180}

\begin{axis}[
height=2.3in,
width=4.5in,
tick align=outside,
tick pos=left,
x grid style={darkgray176},
xlabel={iteration number},
xmin=-0.6, xmax=12.6,
xtick style={color=black},
y grid style={darkgray176},
ylabel={PWM signal \%},
ymin=0, ymax=70,
ytick style={color=black},
xlabel style={font=\fontsize{16}{14}\selectfont},
ylabel style={font=\fontsize{16}{14}\selectfont},
title style={font=\fontsize{16}{14}\selectfont},
title={parabolic distribution},
]
\path [draw=mycolor1, fill=mycolor1, opacity=0.5]
(axis cs:0,50)
--(axis cs:0,50)
--(axis cs:1,43.2958954174903)
--(axis cs:2,50.9434169469502)
--(axis cs:3,51.0649095560523)
--(axis cs:4,52.2041443690175)
--(axis cs:5,53.0094766253504)
--(axis cs:6,53.3782959283732)
--(axis cs:7,53.3912365919861)
--(axis cs:8,53.5587502311926)
--(axis cs:9,54.169454151178)
--(axis cs:10,54.5143896761816)
--(axis cs:11,55.0862617533677)
--(axis cs:12,55.9995823061207)
--(axis cs:12,57.1525427345694)
--(axis cs:12,57.1525427345694)
--(axis cs:11,56.733849839243)
--(axis cs:10,56.2371616340398)
--(axis cs:9,56.2014564933533)
--(axis cs:8,55.5848502326746)
--(axis cs:7,56.2310676886129)
--(axis cs:6,55.2426757513143)
--(axis cs:5,54.3651520774491)
--(axis cs:4,54.7984572545176)
--(axis cs:3,53.9362323099959)
--(axis cs:2,52.6840320886944)
--(axis cs:1,46.0899535815331)
--(axis cs:0,50)
--cycle;

\path [draw=mycolor2, fill=mycolor2, opacity=0.5]
(axis cs:0,50)
--(axis cs:0,50)
--(axis cs:1,53.3598918075788)
--(axis cs:2,57.6405599529536)
--(axis cs:3,64.970351569385)
--(axis cs:4,65)
--(axis cs:5,65)
--(axis cs:6,65)
--(axis cs:7,65)
--(axis cs:8,65)
--(axis cs:9,65)
--(axis cs:10,64.6236018456654)
--(axis cs:11,65)
--(axis cs:12,64.9118344453032)
--(axis cs:12,65.0151268177177)
--(axis cs:12,65.0151268177177)
--(axis cs:11,65)
--(axis cs:10,65.0645797135794)
--(axis cs:9,65)
--(axis cs:8,65)
--(axis cs:7,65)
--(axis cs:6,65)
--(axis cs:5,65)
--(axis cs:4,65)
--(axis cs:3,65.0050868664874)
--(axis cs:2,64.319309431812)
--(axis cs:1,59.7449106419654)
--(axis cs:0,50)
--cycle;

\path [draw=mycolor3, fill=mycolor3, opacity=0.5]
(axis cs:0,50)
--(axis cs:0,50)
--(axis cs:1,34.2868566685366)
--(axis cs:2,43.4788764824336)
--(axis cs:3,44.9590190311818)
--(axis cs:4,48.2352582403966)
--(axis cs:5,51.1200444914652)
--(axis cs:6,52.7870075413554)
--(axis cs:7,54.1005884513405)
--(axis cs:8,54.3999087066124)
--(axis cs:9,53.9179246082147)
--(axis cs:10,53.7367819077438)
--(axis cs:11,53.8986036176419)
--(axis cs:12,53.6985150817721)
--(axis cs:12,57.3082928812813)
--(axis cs:12,57.3082928812813)
--(axis cs:11,57.2431759640639)
--(axis cs:10,56.7677626682653)
--(axis cs:9,57.0720529105027)
--(axis cs:8,57.0958747813433)
--(axis cs:7,56.2125625903261)
--(axis cs:6,55.8413722804219)
--(axis cs:5,56.2401824575908)
--(axis cs:4,57.1416211337896)
--(axis cs:3,58.1573572098664)
--(axis cs:2,56.5029223253464)
--(axis cs:1,46.0883900152199)
--(axis cs:0,50)
--cycle;

\path [draw=mycolor4, fill=mycolor4, opacity=0.5]
(axis cs:0,50)
--(axis cs:0,50)
--(axis cs:1,51.176666608235)
--(axis cs:2,50.9158676770913)
--(axis cs:3,52.7839240543248)
--(axis cs:4,51.8198727612231)
--(axis cs:5,53.1157573480904)
--(axis cs:6,53.4915991849637)
--(axis cs:7,52.9560533053785)
--(axis cs:8,53.7801293158936)
--(axis cs:9,53.7553184171999)
--(axis cs:10,53.6275944298283)
--(axis cs:11,53.5135727139532)
--(axis cs:12,53.5642809312977)
--(axis cs:12,61.7722796359224)
--(axis cs:12,61.7722796359224)
--(axis cs:11,61.7629928378047)
--(axis cs:10,61.6625361854061)
--(axis cs:9,60.2666771272337)
--(axis cs:8,60.4855567192627)
--(axis cs:7,60.1389809760343)
--(axis cs:6,59.6072243623996)
--(axis cs:5,59.7066252609591)
--(axis cs:4,58.2768705045646)
--(axis cs:3,55.992028602739)
--(axis cs:2,56.4309900296145)
--(axis cs:1,55.2982479425463)
--(axis cs:0,50)
--cycle;

\path [draw=mycolor5, fill=mycolor5, opacity=0.5]
(axis cs:0,50)
--(axis cs:0,50)
--(axis cs:1,44.8217664025076)
--(axis cs:2,38.6601823720844)
--(axis cs:3,42.5936441205904)
--(axis cs:4,43.1364460760187)
--(axis cs:5,43.368005080569)
--(axis cs:6,42.9672765418766)
--(axis cs:7,41.057704737538)
--(axis cs:8,40.6162680411954)
--(axis cs:9,42.0481924927773)
--(axis cs:10,43.1472226562644)
--(axis cs:11,42.7147331182825)
--(axis cs:12,43.355505807421)
--(axis cs:12,51.7288803823577)
--(axis cs:12,51.7288803823577)
--(axis cs:11,51.5288531040483)
--(axis cs:10,51.1367408650572)
--(axis cs:9,50.8171868407188)
--(axis cs:8,49.7331785733879)
--(axis cs:7,49.1021488798123)
--(axis cs:6,50.257485421014)
--(axis cs:5,51.1520780289688)
--(axis cs:4,52.0545634454657)
--(axis cs:3,53.994582197769)
--(axis cs:2,52.1181175953635)
--(axis cs:1,49.5796669064116)
--(axis cs:0,50)
--cycle;

\path [draw=mycolor6, fill=mycolor6, opacity=0.5]
(axis cs:0,50)
--(axis cs:0,50)
--(axis cs:1,49.4888236374644)
--(axis cs:2,44.2139281998155)
--(axis cs:3,47.4061597231546)
--(axis cs:4,48.5828109619162)
--(axis cs:5,48.3123627251202)
--(axis cs:6,47.9150207096087)
--(axis cs:7,47.9669823016652)
--(axis cs:8,49.0943064239662)
--(axis cs:9,49.9918552063933)
--(axis cs:10,50.1005006503333)
--(axis cs:11,50.1203672939644)
--(axis cs:12,50.2410396216667)
--(axis cs:12,51.0440098168099)
--(axis cs:12,51.0440098168099)
--(axis cs:11,51.129864131003)
--(axis cs:10,51.7088611253829)
--(axis cs:9,51.6113552428254)
--(axis cs:8,52.0869781626223)
--(axis cs:7,52.1958722108673)
--(axis cs:6,51.9672068701439)
--(axis cs:5,53.0498249783303)
--(axis cs:4,53.4151392740864)
--(axis cs:3,54.4829317685446)
--(axis cs:2,55.3851725488824)
--(axis cs:1,55.5338814406606)
--(axis cs:0,50)
--cycle;

\path [draw=orchid227119194, fill=orchid227119194, opacity=0.5]
(axis cs:0,50)
--(axis cs:0,50)
--(axis cs:1,22.8161821288076)
--(axis cs:2,0)
--(axis cs:3,0)
--(axis cs:4,0)
--(axis cs:5,0)
--(axis cs:6,0)
--(axis cs:7,0)
--(axis cs:8,0)
--(axis cs:9,0)
--(axis cs:10,0)
--(axis cs:11,0)
--(axis cs:12,0)
--(axis cs:12,0)
--(axis cs:12,0)
--(axis cs:11,0)
--(axis cs:10,0)
--(axis cs:9,0)
--(axis cs:8,0)
--(axis cs:7,0)
--(axis cs:6,0)
--(axis cs:5,0)
--(axis cs:4,0)
--(axis cs:3,0)
--(axis cs:2,0)
--(axis cs:1,34.0306119996104)
--(axis cs:0,50)
--cycle;

\addplot [mycolor1, forget plot, line width=\PlotThickness]
table {%
0 50
1 44.6929244995117
2 51.8137245178223
3 52.5005709330241
4 53.5013008117676
5 53.6873143513997
6 54.3104858398438
7 54.8111521402995
8 54.5718002319336
9 55.1854553222656
10 55.3757756551107
11 55.9100557963053
12 56.5760625203451
};
\addplot [mycolor2, forget plot, line width=\PlotThickness]
table {%
0 50
1 56.5524012247721
2 60.9799346923828
3 64.9877192179362
4 65
5 65
6 65
7 65
8 65
9 65
10 64.8440907796224
11 65
12 64.9634806315104
};
\addplot [mycolor3, forget plot, line width=\PlotThickness]
table {%
0 50
1 40.1876233418783
2 49.99089940389
3 51.5581881205241
4 52.6884396870931
5 53.680113474528
6 54.3141899108887
7 55.1565755208333
8 55.7478917439779
9 55.4949887593587
10 55.2522722880046
11 55.5708897908529
12 55.5034039815267
};
\addplot [semithick, mycolor4, forget plot, line width=\PlotThickness]
table {%
0 50
1 53.2374572753906
2 53.6734288533529
3 54.3879763285319
4 55.0483716328939
5 56.4111913045247
6 56.5494117736816
7 56.5475171407064
8 57.1328430175781
9 57.0109977722168
10 57.6450653076172
11 57.6382827758789
12 57.66828028361
};
\addplot [semithick, mycolor5, forget plot, line width=\PlotThickness]
table {%
0 50
1 47.2007166544596
2 45.389149983724
3 48.2941131591797
4 47.5955047607422
5 47.2600415547689
6 46.6123809814453
7 45.0799268086751
8 45.1747233072917
9 46.432689666748
10 47.1419817606608
11 47.1217931111654
12 47.5421930948893
};
\addplot [semithick, mycolor6, forget plot, line width=\PlotThickness]
table {%
0 50
1 52.5113525390625
2 49.799550374349
3 50.9445457458496
4 50.9989751180013
5 50.6810938517253
6 49.9411137898763
7 50.0814272562663
8 50.5906422932943
9 50.8016052246094
10 50.9046808878581
11 50.6251157124837
12 50.6425247192383
};

\addplot [semithick, orchid227119194, forget plot, line width=\PlotThickness]
table {%
0 50
1 28.423397064209
2 0
3 0
4 0
5 0
6 0
7 0
8 0
9 0
10 0
11 0
12 0
};
\end{axis}

\end{tikzpicture}%
    \label{fig:pwm_uniform_fill}}
    \hfill
    \subfloat[]{
\begin{tikzpicture}[scale = 0.5]

\definecolor{cyan}{RGB}{0,255,255}
\definecolor{darkgray176}{RGB}{176,176,176}
\definecolor{lightgray204}{RGB}{204,204,204}
\definecolor{magenta}{RGB}{255,0,255}
\definecolor{sandybrown}{RGB}{244,164,96}
\definecolor{orchid227119194}{RGB}{227,119,194}

\begin{axis}[
height=2.3in,
width=4.5in,
legend cell align={left},
legend style={
  fill opacity=0.8,
  draw opacity=1,
  text opacity=1,
  at={(0.97,0.03)},
  anchor=south east,
  draw=lightgray204
},
tick align=outside,
tick pos=left,
x grid style={darkgray176},
xlabel={iteration number},
xmin=-0.6, xmax=12.6,
xtick style={color=black},
y grid style={darkgray176}, 
ymin=0, ymax=70,
ytick style={color=black},
yticklabels=\empty,
xlabel style={font=\fontsize{16}{14}\selectfont},
ylabel style={font=\fontsize{16}{14}\selectfont},
title style={font=\fontsize{16}{14}\selectfont},
title={Gaussian distribution},
]
\path [draw=mycolor1, fill=mycolor1, opacity=0.5]
(axis cs:0,50)
--(axis cs:0,50)
--(axis cs:1,51.1233596801758)
--(axis cs:2,53.2094612121582)
--(axis cs:3,53.2648811340332)
--(axis cs:4,55.360050201416)
--(axis cs:5,55.4710998535156)
--(axis cs:6,55.6345977783203)
--(axis cs:7,55.848575592041)
--(axis cs:8,56.9167022705078)
--(axis cs:9,56.4805335998535)
--(axis cs:10,57.0560302734375)
--(axis cs:11,57.2796173095703)
--(axis cs:12,57.3451271057129)
--(axis cs:12,58.444034576416)
--(axis cs:12,58.444034576416)
--(axis cs:11,57.8418655395508)
--(axis cs:10,57.432975769043)
--(axis cs:9,56.8183937072754)
--(axis cs:8,57.1423721313477)
--(axis cs:7,56.535888671875)
--(axis cs:6,56.5028495788574)
--(axis cs:5,56.569019317627)
--(axis cs:4,55.5849571228027)
--(axis cs:3,56.4828567504883)
--(axis cs:2,56.6209411621094)
--(axis cs:1,53.9216194152832)
--(axis cs:0,50)
--cycle;

\path [draw=mycolor2, fill=mycolor2, opacity=0.5]
(axis cs:0,50)
--(axis cs:0,50)
--(axis cs:1,32.9232063293457)
--(axis cs:2,18.3353805541992)
--(axis cs:3,17.3033313751221)
--(axis cs:4,15.4551076889038)
--(axis cs:5,19.9672260284424)
--(axis cs:6,19.1648349761963)
--(axis cs:7,13.4675416946411)
--(axis cs:8,10.755952835083)
--(axis cs:9,3.21574378013611)
--(axis cs:10,6.07384252548218)
--(axis cs:11,8.51219940185547)
--(axis cs:12,6.11319017410278)
--(axis cs:12,24.2602024078369)
--(axis cs:12,24.2602024078369)
--(axis cs:11,24.0177440643311)
--(axis cs:10,23.1807918548584)
--(axis cs:9,23.0534973144531)
--(axis cs:8,23.3352966308594)
--(axis cs:7,23.3955173492432)
--(axis cs:6,22.7124671936035)
--(axis cs:5,22.7942028045654)
--(axis cs:4,22.3358364105225)
--(axis cs:3,25.561824798584)
--(axis cs:2,28.3551235198975)
--(axis cs:1,42.0860862731934)
--(axis cs:0,50)
--cycle;

\path [draw=mycolor3, fill=mycolor3, opacity=0.5]
(axis cs:0,50)
--(axis cs:0,50)
--(axis cs:1,25.8157405853271)
--(axis cs:2,22.8432960510254)
--(axis cs:3,18.5912666320801)
--(axis cs:4,18.950927734375)
--(axis cs:5,19.6618309020996)
--(axis cs:6,18.056489944458)
--(axis cs:7,18.1301670074463)
--(axis cs:8,18.2613105773926)
--(axis cs:9,17.7889251708984)
--(axis cs:10,17.831226348877)
--(axis cs:11,16.6421699523926)
--(axis cs:12,16.3682518005371)
--(axis cs:12,19.0040493011475)
--(axis cs:12,19.0040493011475)
--(axis cs:11,18.4329643249512)
--(axis cs:10,18.4927825927734)
--(axis cs:9,18.3100090026855)
--(axis cs:8,18.2858715057373)
--(axis cs:7,20.1170024871826)
--(axis cs:6,23.113187789917)
--(axis cs:5,27.1687602996826)
--(axis cs:4,28.1219348907471)
--(axis cs:3,28.5082054138184)
--(axis cs:2,25.9272899627686)
--(axis cs:1,27.4734745025635)
--(axis cs:0,50)
--cycle;

\path [draw=mycolor4, fill=mycolor4, opacity=0.5]
(axis cs:0,50)
--(axis cs:0,50)
--(axis cs:1,41.3378715515137)
--(axis cs:2,32.0553512573242)
--(axis cs:3,33.5796241760254)
--(axis cs:4,33.3937568664551)
--(axis cs:5,31.4837894439697)
--(axis cs:6,31.890323638916)
--(axis cs:7,32.8692436218262)
--(axis cs:8,31.60866355896)
--(axis cs:9,32.5548400878906)
--(axis cs:10,32.3891448974609)
--(axis cs:11,31.9480476379395)
--(axis cs:12,31.3908767700195)
--(axis cs:12,35.965763092041)
--(axis cs:12,35.965763092041)
--(axis cs:11,35.533878326416)
--(axis cs:10,36.3164825439453)
--(axis cs:9,36.7598991394043)
--(axis cs:8,34.1019859313965)
--(axis cs:7,34.6838073730469)
--(axis cs:6,34.1934471130371)
--(axis cs:5,34.2432708740234)
--(axis cs:4,36.8381042480469)
--(axis cs:3,36.8839950561523)
--(axis cs:2,39.4886627197266)
--(axis cs:1,44.1568489074707)
--(axis cs:0,50)
--cycle;

\path [draw=mycolor5, fill=mycolor5, opacity=0.5]
(axis cs:0,50)
--(axis cs:0,50)
--(axis cs:1,20.4195022583008)
--(axis cs:2,12.8993091583252)
--(axis cs:3,2.97156286239624)
--(axis cs:4,0)
--(axis cs:5,0)
--(axis cs:6,6.34952116012573)
--(axis cs:7,4.81620311737061)
--(axis cs:8,3.97523713111877)
--(axis cs:9,2.39970660209656)
--(axis cs:10,0)
--(axis cs:11,0)
--(axis cs:12,0)
--(axis cs:12,4.20497131347656)
--(axis cs:12,4.20497131347656)
--(axis cs:11,6.81052017211914)
--(axis cs:10,8.91381072998047)
--(axis cs:9,9.13401031494141)
--(axis cs:8,9.93240356445312)
--(axis cs:7,9.6885871887207)
--(axis cs:6,13.8590965270996)
--(axis cs:5,6.56165170669556)
--(axis cs:4,11.5486450195312)
--(axis cs:3,14.8001184463501)
--(axis cs:2,15.0162830352783)
--(axis cs:1,24.1283740997314)
--(axis cs:0,50)
--cycle;

\path [draw=mycolor6, fill=mycolor6, opacity=0.5]
(axis cs:0,50)
--(axis cs:0,50)
--(axis cs:1,41.827320098877)
--(axis cs:2,28.2066135406494)
--(axis cs:3,29.8571491241455)
--(axis cs:4,28.0124588012695)
--(axis cs:5,30.353588104248)
--(axis cs:6,30.6691970825195)
--(axis cs:7,29.7522315979004)
--(axis cs:8,28.2299098968506)
--(axis cs:9,27.7879962921143)
--(axis cs:10,28.0488624572754)
--(axis cs:11,28.3183898925781)
--(axis cs:12,28.2041816711426)
--(axis cs:12,34.9073524475098)
--(axis cs:12,34.9073524475098)
--(axis cs:11,35.566219329834)
--(axis cs:10,36.5052490234375)
--(axis cs:9,37.0212326049805)
--(axis cs:8,37.0407333374023)
--(axis cs:7,36.661003112793)
--(axis cs:6,32.9619102478027)
--(axis cs:5,32.7368583679199)
--(axis cs:4,34.0292701721191)
--(axis cs:3,31.7350673675537)
--(axis cs:2,37.9810523986816)
--(axis cs:1,42.4186477661133)
--(axis cs:0,50)
--cycle;

\path [draw=orchid227119194, fill=orchid227119194, opacity=0.5]
(axis cs:0,50)
--(axis cs:0,50)
--(axis cs:1,40.8203620910645)
--(axis cs:2,34.0650901794434)
--(axis cs:3,33.5649223327637)
--(axis cs:4,31.1327648162842)
--(axis cs:5,29.4384784698486)
--(axis cs:6,32.6307945251465)
--(axis cs:7,29.2737140655518)
--(axis cs:8,29.3416862487793)
--(axis cs:9,27.2634162902832)
--(axis cs:10,26.2653045654297)
--(axis cs:11,27.4140090942383)
--(axis cs:12,27.9772243499756)
--(axis cs:12,38.5415496826172)
--(axis cs:12,38.5415496826172)
--(axis cs:11,38.2241096496582)
--(axis cs:10,37.9177742004395)
--(axis cs:9,38.021900177002)
--(axis cs:8,37.9864196777344)
--(axis cs:7,38.6219520568848)
--(axis cs:6,37.8175239562988)
--(axis cs:5,38.5721969604492)
--(axis cs:4,39.7640190124512)
--(axis cs:3,42.2031898498535)
--(axis cs:2,44.7263717651367)
--(axis cs:1,50.5758934020996)
--(axis cs:0,50)
--cycle;

\addplot [semithick, mycolor1, forget plot, line width=\PlotThickness]
table {%
0 50
1 52.5224895477295
2 54.9152011871338
3 54.8738689422607
4 55.4725036621094
5 56.0200595855713
6 56.0687236785889
7 56.192232131958
8 57.0295372009277
9 56.6494636535645
10 57.2445030212402
11 57.5607414245605
12 57.8945808410645
};
\addplot [semithick, mycolor2, forget plot, line width=\PlotThickness]
table {%
0 50
1 37.5046463012695
2 23.3452520370483
3 21.432578086853
4 18.8954720497131
5 21.3807144165039
6 20.9386510848999
7 18.4315295219421
8 17.0456247329712
9 13.1346205472946
10 14.6273171901703
11 16.2649717330933
12 15.1866962909698
};
\addplot [semithick, mycolor3, forget plot, line width=\PlotThickness]
table {%
0 50
1 26.6446075439453
2 24.385293006897
3 23.5497360229492
4 23.536431312561
5 23.4152956008911
6 20.5848388671875
7 19.1235847473145
8 18.2735910415649
9 18.049467086792
10 18.1620044708252
11 17.5375671386719
12 17.6861505508423
};
\addplot [semithick, mycolor4, forget plot, line width=\PlotThickness]
table {%
0 50
1 42.7473602294922
2 35.7720069885254
3 35.2318096160889
4 35.115930557251
5 32.8635301589966
6 33.0418853759766
7 33.7765254974365
8 32.8553247451782
9 34.6573696136475
10 34.3528137207031
11 33.7409629821777
12 33.6783199310303
};
\addplot [semithick, mycolor5, forget plot, line width=\PlotThickness]
table {%
0 50
1 22.2739381790161
2 13.9577960968018
3 8.88584065437317
4 5.77432250976562
5 3.28082585334778
6 10.1043088436127
7 7.25239515304565
8 6.95382034778595
9 5.76685845851898
10 4.45690536499024
11 3.40526008605957
12 2.10248565673828
};
\addplot [semithick, mycolor6, forget plot, line width=\PlotThickness]
table {%
0 50
1 42.1229839324951
2 33.0938329696655
3 30.7961082458496
4 31.0208644866943
5 31.545223236084
6 31.8155536651611
7 33.2066173553467
8 32.6353216171265
9 32.4046144485474
10 32.2770557403564
11 31.9423046112061
12 31.5557670593262
};
\addplot [semithick, orchid227119194, forget plot, line width=\PlotThickness]
table {%
0 50
1 45.698127746582
2 39.39573097229
3 37.8840560913086
4 35.4483919143677
5 34.0053377151489
6 35.2241592407227
7 33.9478330612183
8 33.6640529632568
9 32.6426582336426
10 32.0915393829346
11 32.8190593719482
12 33.2593870162964
};
\end{axis}

\end{tikzpicture}%
    \label{fig:pwm_Gaussian_fill}}
    \hfill
    \subfloat[]{\input{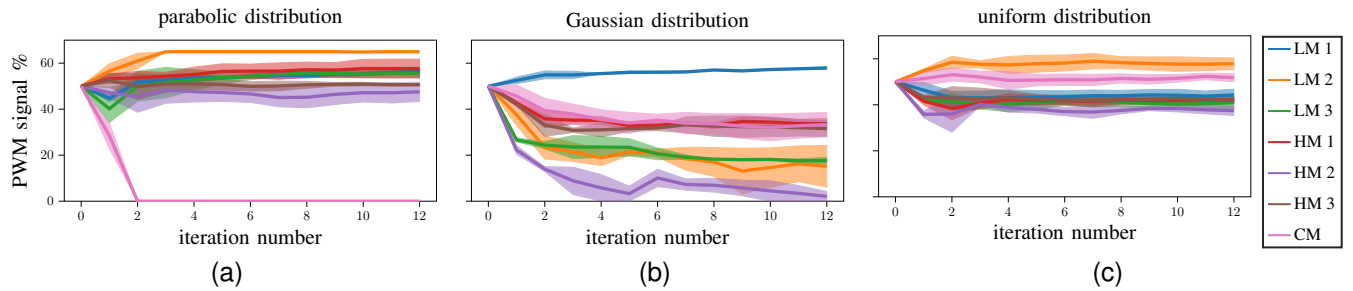}%
    \label{fig:pwm_linear_fill}}
    \hfill
    \caption{This figure demonstrates the motors' PWM commands. Plot (a) shows the commands for a parabolic target distribution. Plot (b) shows the motor commands for a Gaussian target distribution, while Plot (c) demonstrates the case for a uniform target distribution. LM refers to a lower motor (clock-wise rotating motor), HM refers to a higher motor (counter clock-wise rotating motor), and CM refers to the central motor. The shaded regions indicate the standard deviation over five experiments.}%
    \label{fig:PWM_signals}%
\end{figure*}

\textbf{Experiment 1 - Fit to a flying distribution:}
In this experiment, we design the desired force field to be suitable for flight with a soaring robot where the airflow at the center is lower than the edges, that is, $F_\mathrm{des}(x,y) = 2.25 + 30({x}^{2}+{y}^{2})$. We also performed twelve iterations and the convergence to the target distribution is shown in the first column of Fig.~\ref{fig:algorithm_performance_all}, and the PWM signals are shown in Fig.~\ref{fig:PWM_signals}~(a). This highlights how the central motor speed converges to zero after two iterations (the speeds are clipped at zero to ensure physical feasibility). The effectiveness of this experiment is illustrated in the supplementary Video~1. Following the adaptation of the airflow, the flight experiment showed a $50\%$ reduction in the average position tracking error per step and a $20\%$ reduction in the average attitude tracking error per step.

\textbf{Experiment 2 - Fit to a Gaussian distribution:}
In this experiment, we set the desired force field to be a Gaussian distribution centered over one of the motors (referred to as \textbf{LM1}). We applied twelve iterations and the convergence to the desired distribution is shown in the 2nd and 4th columns of Fig.~\ref{fig:algorithm_performance_all}; Fig.~\ref{fig:PWM_signals} (b) shows the PWM signals where the motor \textbf{LM1} has the highest PWM signal as expected. We disabled the central motor and our algorithm was still able to achieve a Gaussian distribution as shown in the 4th column of Fig.~\ref{fig:algorithm_performance_all}.

\textbf{Experiment 3 - Fit to a uniform distribution:}
In this experiment, we set the desired force field to be uniform with value $F_\mathrm{des} = 2.5\ \mathrm{N}$. We performed our algorithm and found that the measured force distribution converges to the desired uniform distribution (see the 3rd and 5th columns of Fig.~\ref{fig:algorithm_performance_all}), and Fig.~\ref{fig:PWM_signals} (c) shows the PWM signals. We disabled the central motor and our algorithm was still able to achieve a uniform distribution as shown in the 5th column of Fig.~\ref{fig:algorithm_performance_all}. We note that the algorithm was tested over 20 iterations (see Appendix \ref{sec:appendix_convergence} and Fig.~\ref{fig:Convergence_PWM_signals}), but no further improvement was observed after six to seven iterations. This shows that the algorithm indeed converges. 

Overall, we observed in all experiments that the PWM signals stabilized and the error value rapidly decayed. However, the error converges to a nonzero value (typically after six iterations), which is due to the fact that the system has many more degrees of freedom (represented by the $317$ key points) than the number of controls (seven motors). From a control perspective, this means that we are dealing with a largely under-actuated system and cannot expect to perfectly match a given distribution except in very special cases. 
Our algorithm showed robustness to changes in the number of used motors, which highlights one of its advantages over numerical simulations, which need to be recalibrated after any minor change to the setup. 

\section{Convergence}
\label{sec:Convergence}
In the following, we will simplify the notation for the sake of readability. We will denote by $F(U)$ the vector $(F(U, \hat{z})_{\hat{z} \in \Omega})$ that contains the force measurements at the key points and by $F_{\text{des}}$ the vector $(F_{\text{des}}(\hat{z})_{\hat{z} \in \Omega})$ that contains the desired forces at the key points. We further introduce the error function $\zeta(U) = \frac{1}{2} \|F(U) - F_{\text{des}}\|_2^2$, where $\|\cdot\|_2$ denotes the Euclidean distance, as this will further simplify the statement of our convergence result.

Due to the under-actuated nature of the system ($m = 7$ control inputs, $N = 317$ target points), the error cannot in general reach zero. Let $\zeta^* = \min_U \zeta(U) > 0$ represent the minimum achievable residual error for a specific desired distribution. 

\textit{Proposition 1:} If there exist constants $L > 0$, $c > 0$, and $\mu > 0$ such that the following three conditions are satisfied for all $U \in \mathbb{R}^{N_{\mathrm{motors}}}$:
\begin{subequations}\label{eq:assumptions}
\begin{align}
\LNorm{\frac{\partial^{2} \zeta}{\partial U^{2}}}_{2} \le& L \; ; \label{eq:assumptionsA}\\
-\frac{1}{2} \frac{\partial \Phi_{\mathrm{pr}}}{\partial U}^{+^\top}\frac{\partial F}{\partial U}^\top - \frac{1}{2}\frac{\partial F}{\partial U}\frac{\partial \Phi_{\mathrm{pr}}}{\partial U}^{+} & \nonumber \\
+ \frac{L}{2}\lambda \frac{\partial \Phi_{\mathrm{pr}}}{\partial U}^{+^\top}\frac{\partial \Phi_{\mathrm{pr}}}{\partial U}^{+} \le& -c \frac{\partial F}{\partial U}\frac{\partial F}{\partial U}^\top ; \label{eq:assumptionsB}\\
\LNorm{\frac{\partial \zeta}{\partial U}}_{2}^{2} \ge& 2\mu (\zeta(U) - \zeta^*) \;,  \label{eq:assumptionsC}
\end{align}
\end{subequations}
where $\lambda > 0$ is the learning rate, then the iterates of Algorithm 1 satisfy:
\begin{equation}
    \zeta(U_k) - \zeta^* \le (1 - 2\mu c \lambda)^k (\zeta(U_0) - \zeta^*), \label{eq:convergence_prop}
\end{equation}
for all $k \ge 0$. Thus, the tracking error $\zeta(U_k)$ converges linearly to the minimum residual error $\zeta^*$, provided that the learning rate is chosen such that $\lambda \in (0, 1 / (2\mu c))$.

The proof of the proposition is included in Appendix II. The three assumptions \eqref{eq:assumptionsA}-\eqref{eq:assumptionsC} have the following interpretation. The condition \eqref{eq:assumptionsA} ensures that the error function $\zeta(U)$ is $L$-smooth. The consolidated condition \eqref{eq:assumptionsB} serves a dual purpose: it guarantees that the search direction generated by our physical model's pseudo-inverse is sufficiently aligned with the true system's gradient, while simultaneously ensuring that the pseudo-inverse remains bounded. This inequality acts as a combined gradient-alignment and step-size condition. Finally, \eqref{eq:assumptionsC} (Polyak-Lojasiewicz) means that as long as our tracking error is higher than $\zeta^*$, the system's gradient remains sufficiently large to decrease the error. We note that a weaker form of convergence holds even if condition \eqref{eq:assumptionsC} is violated.


\section{Conclusion}
\label{sec:conclusion}
In this work, we presented an online learning algorithm to solve the challenge of precisely shaping the complex airflow in a multi-fan wind tunnel, a critical capability for the experimental testing of advanced aerial robots. Our method successfully combines a simplified physical model with iterative, measurement-based learning to achieve sample-efficient convergence to a variety of target airflow distributions.
We experimentally demonstrated the algorithm's versatility by generating uniform, Gaussian, and complex parabolic profiles. Critically, we showed that our method can produce a specialized airflow profile designed for passive soaring, improving the flight performance of a soaring robot. Furthermore, we highlighted the algorithm's robustness to significant system changes, such as the deactivation of a motor, underscoring its practical utility for real-world experimental testbeds. While our current setup utilizes a moving sensor as shown in supplementary Video~1, the high sample efficiency of our algorithm ensures convergence in only six to seven iterations, allowing the entire process to be completed in a few minutes, with each iteration taking about $20\,\mathrm{s}$ (see supplementary Video~1). For real-time scalability, this moving sensor could easily be replaced by a fixed grid of static sensors to provide instantaneous spatial measurements at each iteration.


\section*{Acknowledgments}
The authors thank the International Max Planck Research School for Intelligent Systems (IMPRS-IS) for supporting Ghadeer Elmkaiel. We also thank the German Research Foundation for the support.

\appendices

\section{Thrust model of the motors} \label{sec:poly_fit}
\begin{figure}
    \vspace{6pt}
    \centering
    \includegraphics[width=3.1in]{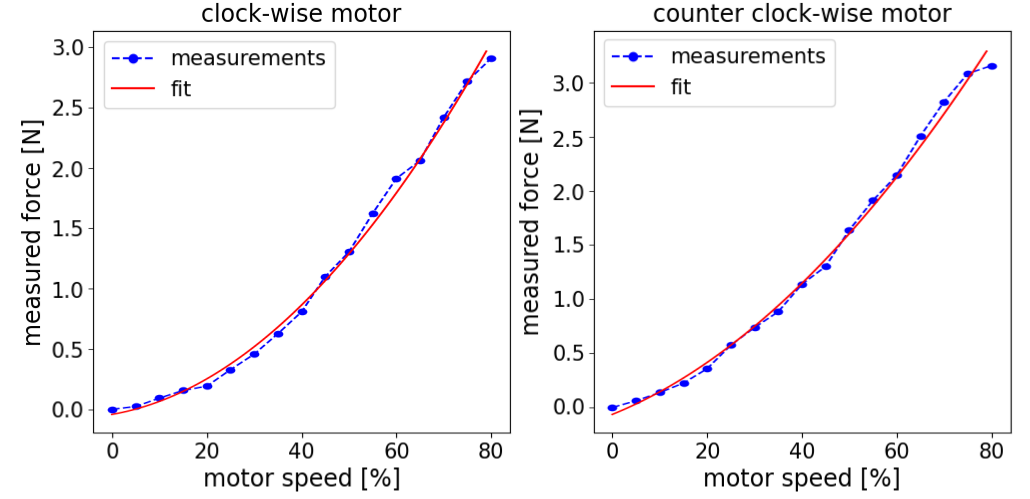}
    \caption{This figure shows the polynomial fit, which maps the PWM commands to the force measurements exactly on top of the motor for two different motors. The plot on the left shows the results for a clock-wise rotating motor, while the plot on the right shows the results for a counter clock-wise rotating motor.}
    \label{fig:polynomial_fit}
\end{figure}

We make the assumption that the function $P_\mathrm{h/l}(U_j)$ in (\ref{eq:prior}), which maps the PWM commands of motor $j$ to the force measured on top of the motor, is a polynomial of degree two. 
As shown in Figure \ref{fig:polynomial_fit}, the resulting fit explains the force measurements well. The force measurements are obtained by changing the PWM signal from $0$ to $80\%$ by increments of $5\%$. We also note that the right panel shows the result for a counter clock-wise rotating motor $P_\mathrm{h}(U_j)$, which is located at a slightly higher position than the clock-wise rotating motor, shown on the left panel $P_\mathrm{l}(U_j)$. This is reflected in our measurements and our fit, where the counter clock-wise rotating motor on the right panel produces a larger force.

\section{Convergence experiments and proof of Prop. 1} \label{sec:appendix_convergence}
\begin{figure*}[!t]
    \centering
    \subfloat[]{
\begin{tikzpicture}[scale = 0.5]

\definecolor{cyan}{RGB}{0,255,255}
\definecolor{darkgray176}{RGB}{176,176,176}
\definecolor{lightgray204}{RGB}{204,204,204}
\definecolor{magenta}{RGB}{255,0,255}
\definecolor{sandybrown}{RGB}{244,164,96}

\begin{axis}[
height=2.3in,
width=4.4in,
tick align=outside,
tick pos=left,
x grid style={darkgray176},
xlabel={iteration number},
xmin=-0.6, xmax=12.6,
xtick style={color=black},
y grid style={darkgray176},
ylabel={PWM signal \%},
ymin=15, ymax=55,
ytick style={color=black},
xlabel style={font=\fontsize{16}{14}\selectfont},
ylabel style={font=\fontsize{16}{14}\selectfont},
title style={font=\fontsize{16}{14}\selectfont},
title={same initial speeds},
]
\path [draw=mycolor1, fill=mycolor1, opacity=0.5]
(axis cs:0,50)
--(axis cs:0,50)
--(axis cs:1,50)
--(axis cs:2,34.1237040391164)
--(axis cs:3,44.7862113473045)
--(axis cs:4,42.6376703166196)
--(axis cs:5,42.3970659160002)
--(axis cs:6,42.6041441920905)
--(axis cs:7,40.671947099573)
--(axis cs:8,41.8076677298635)
--(axis cs:9,41.4457813151671)
--(axis cs:10,41.7193571932072)
--(axis cs:11,41.9334898374426)
--(axis cs:12,42.5710144844785)
--(axis cs:12,44.2969855155216)
--(axis cs:12,44.2969855155216)
--(axis cs:11,44.2705101625574)
--(axis cs:10,44.2526428067928)
--(axis cs:9,44.0982186848329)
--(axis cs:8,44.4203322701365)
--(axis cs:7,45.360052900427)
--(axis cs:6,46.0438558079095)
--(axis cs:5,47.6309340839998)
--(axis cs:4,47.7823296833804)
--(axis cs:3,48.1737886526955)
--(axis cs:2,44.5442959608836)
--(axis cs:1,50)
--(axis cs:0,50)
--cycle;

\path [draw=mycolor2, fill=mycolor2, opacity=0.5]
(axis cs:0,50)
--(axis cs:0,50)
--(axis cs:1,50)
--(axis cs:2,28.6525050180395)
--(axis cs:3,44.5072382502539)
--(axis cs:4,43.6022692074664)
--(axis cs:5,43.7228613199755)
--(axis cs:6,41.9221212605935)
--(axis cs:7,39.4146412773715)
--(axis cs:8,40.6556909332127)
--(axis cs:9,40.8340495059616)
--(axis cs:10,41.001106037344)
--(axis cs:11,41.4910937961175)
--(axis cs:12,41.6803017395578)
--(axis cs:12,43.1116982604422)
--(axis cs:12,43.1116982604422)
--(axis cs:11,43.0369062038825)
--(axis cs:10,43.122893962656)
--(axis cs:9,43.2579504940384)
--(axis cs:8,44.6163090667873)
--(axis cs:7,45.1213587226285)
--(axis cs:6,44.7258787394065)
--(axis cs:5,45.6771386800245)
--(axis cs:4,47.0937307925336)
--(axis cs:3,49.480761749746)
--(axis cs:2,39.4634949819605)
--(axis cs:1,50)
--(axis cs:0,50)
--cycle;

\path [draw=mycolor3, fill=mycolor3, opacity=0.5]
(axis cs:0,50)
--(axis cs:0,50)
--(axis cs:1,50)
--(axis cs:2,38.3167839912873)
--(axis cs:3,46.4943254083217)
--(axis cs:4,43.8168878581738)
--(axis cs:5,44.8984252370362)
--(axis cs:6,45.0019284863982)
--(axis cs:7,43.7453184141121)
--(axis cs:8,45.2080140845705)
--(axis cs:9,44.8637564129659)
--(axis cs:10,45.0731412398931)
--(axis cs:11,45.4867457354148)
--(axis cs:12,45.4528167917363)
--(axis cs:12,47.5871832082637)
--(axis cs:12,47.5871832082637)
--(axis cs:11,47.4652542645852)
--(axis cs:10,47.7988587601069)
--(axis cs:9,47.8362435870341)
--(axis cs:8,48.3319859154294)
--(axis cs:7,48.0946815858879)
--(axis cs:6,48.3020715136018)
--(axis cs:5,48.6135747629638)
--(axis cs:4,49.0871121418262)
--(axis cs:3,50.1816745916783)
--(axis cs:2,48.5712160087127)
--(axis cs:1,50)
--(axis cs:0,50)
--cycle;

\path [draw=mycolor4, fill=mycolor4, opacity=0.5]
(axis cs:0,50)
--(axis cs:0,50)
--(axis cs:1,28.7314405935424)
--(axis cs:2,37.1272685415995)
--(axis cs:3,29.4228071068564)
--(axis cs:4,27.3542056359094)
--(axis cs:5,27.5042923770384)
--(axis cs:6,30.5252946423125)
--(axis cs:7,30.7864254928094)
--(axis cs:8,30.9249830937205)
--(axis cs:9,31.3797747266507)
--(axis cs:10,31.0018801400613)
--(axis cs:11,31.1942593621939)
--(axis cs:12,30.80249264077)
--(axis cs:12,34.30150735923)
--(axis cs:12,34.30150735923)
--(axis cs:11,34.7257406378061)
--(axis cs:10,35.6741198599387)
--(axis cs:9,36.3162252733493)
--(axis cs:8,36.0710169062795)
--(axis cs:7,37.0375745071906)
--(axis cs:6,35.0627053576875)
--(axis cs:5,35.8357076229616)
--(axis cs:4,36.7297943640906)
--(axis cs:3,35.9291928931436)
--(axis cs:2,49.0767314584005)
--(axis cs:1,36.2685594064576)
--(axis cs:0,50)
--cycle;

\path [draw=mycolor5, fill=mycolor5, opacity=0.5]
(axis cs:0,50)
--(axis cs:0,50)
--(axis cs:1,24.8882162538281)
--(axis cs:2,37.7644223865457)
--(axis cs:3,29.9591282276099)
--(axis cs:4,30.7737234849451)
--(axis cs:5,31.1104627038973)
--(axis cs:6,33.020172815108)
--(axis cs:7,32.8375252490627)
--(axis cs:8,32.251417232136)
--(axis cs:9,32.6466153220458)
--(axis cs:10,32.0139508810039)
--(axis cs:11,31.9460434787501)
--(axis cs:12,31.1416796639528)
--(axis cs:12,36.0743203360472)
--(axis cs:12,36.0743203360472)
--(axis cs:11,35.8099565212499)
--(axis cs:10,36.3300491189961)
--(axis cs:9,36.4293846779542)
--(axis cs:8,35.800582767864)
--(axis cs:7,37.7584747509373)
--(axis cs:6,35.843827184892)
--(axis cs:5,36.5735372961027)
--(axis cs:4,38.0502765150549)
--(axis cs:3,36.3928717723901)
--(axis cs:2,53.9955776134543)
--(axis cs:1,31.879783746172)
--(axis cs:0,50)
--cycle;

\path [draw=mycolor6, fill=mycolor6, opacity=0.5]
(axis cs:0,50)
--(axis cs:0,50)
--(axis cs:1,24.4855709085016)
--(axis cs:2,21.3868181792088)
--(axis cs:3,24.1940725721312)
--(axis cs:4,25.6082214926714)
--(axis cs:5,24.9824554188938)
--(axis cs:6,25.654364011092)
--(axis cs:7,27.3568623730045)
--(axis cs:8,27.0065561834051)
--(axis cs:9,27.0578361460276)
--(axis cs:10,26.5800702398504)
--(axis cs:11,26.9924585711685)
--(axis cs:12,26.1764256135278)
--(axis cs:12,31.5235743864722)
--(axis cs:12,31.5235743864722)
--(axis cs:11,31.3795414288315)
--(axis cs:10,32.1039297601496)
--(axis cs:9,32.2821638539724)
--(axis cs:8,31.7454438165949)
--(axis cs:7,32.5231376269955)
--(axis cs:6,31.829635988908)
--(axis cs:5,33.5455445811062)
--(axis cs:4,35.2877785073286)
--(axis cs:3,34.9419274278687)
--(axis cs:2,43.5571818207912)
--(axis cs:1,32.9224290914984)
--(axis cs:0,50)
--cycle;

\addplot [mycolor1, forget plot, line width=\PlotThickness]
table {%
0 50
1 50
2 39.334
3 46.48
4 45.21
5 45.014
6 44.324
7 43.016
8 43.114
9 42.772
10 42.986
11 43.102
12 43.434
};
\addplot [mycolor2, forget plot, line width=\PlotThickness]
table {%
0 50
1 50
2 34.058
3 46.994
4 45.348
5 44.7
6 43.324
7 42.268
8 42.636
9 42.046
10 42.062
11 42.264
12 42.396
};
\addplot [mycolor3, forget plot, line width=\PlotThickness]
table {%
0 50
1 50
2 43.444
3 48.338
4 46.452
5 46.756
6 46.652
7 45.92
8 46.77
9 46.35
10 46.436
11 46.476
12 46.52
};
\addplot [semithick, mycolor4, forget plot, line width=\PlotThickness]
table {%
0 50
1 32.5
2 43.102
3 32.676
4 32.042
5 31.67
6 32.794
7 33.912
8 33.498
9 33.848
10 33.338
11 32.96
12 32.552
};
\addplot [semithick, mycolor5, forget plot, line width=\PlotThickness]
table {%
0 50
1 28.384
2 45.88
3 33.176
4 34.412
5 33.842
6 34.432
7 35.298
8 34.026
9 34.538
10 34.172
11 33.878
12 33.608
};
\addplot [semithick, mycolor6, forget plot, line width=\PlotThickness]
table {%
0 50
1 28.704
2 32.472
3 29.568
4 30.448
5 29.264
6 28.742
7 29.94
8 29.376
9 29.67
10 29.342
11 29.186
12 28.85
};
\end{axis}

\end{tikzpicture}%
    \label{fig:pwm_uniform_fill_appendix}}
    \hfill
    \subfloat[]{
\begin{tikzpicture}[scale = 0.5]


\begin{axis}[
height=2.3in,
width=4.4in,
legend cell align={left},
legend style={
  fill opacity=0.8,
  draw opacity=1,
  text opacity=1,
  at={(0.97,0.03)},
  anchor=south east,
  draw=lightgray204
},
tick align=outside,
tick pos=left,
x grid style={darkgray176},
xlabel={iteration number},
xmin=-0.6, xmax=12.6,
xtick style={color=black},
y grid style={darkgray176}, 
ymin=15, ymax=55,
ytick style={color=black},
yticklabels=\empty,
legend style={legend cell align=left, align=left, draw=white!15!black, font=\tiny, legend pos=outer north east},
xlabel style={font=\fontsize{16}{14}\selectfont},
ylabel style={font=\fontsize{16}{14}\selectfont},
title style={font=\fontsize{16}{14}\selectfont},
title={different initial speeds},
]
\path [draw=mycolor1, fill=mycolor1, opacity=0.5]
(axis cs:0,47.9664983122039)
--(axis cs:0,41.3668350211294)
--(axis cs:1,42.8896465982893)
--(axis cs:2,37.6447199467439)
--(axis cs:3,38.5511140381709)
--(axis cs:4,42.988860255808)
--(axis cs:5,41.035372499064)
--(axis cs:6,42.2538107448188)
--(axis cs:7,41.7200287367)
--(axis cs:8,41.6408241451428)
--(axis cs:9,40.7852827177766)
--(axis cs:10,40.8165781397434)
--(axis cs:11,41.598383683634)
--(axis cs:12,41.7786770339557)
--(axis cs:12,44.307989632711)
--(axis cs:12,44.307989632711)
--(axis cs:11,44.301616316366)
--(axis cs:10,43.7034218602566)
--(axis cs:9,44.1747172822234)
--(axis cs:8,43.3791758548572)
--(axis cs:7,42.4933045966333)
--(axis cs:6,42.4395225885146)
--(axis cs:5,44.1446275009361)
--(axis cs:4,43.2044730775253)
--(axis cs:3,42.7688859618291)
--(axis cs:2,44.9352800532561)
--(axis cs:1,48.863686735044)
--(axis cs:0,47.9664983122039)
--cycle;

\path [draw=mycolor2, fill=mycolor2, opacity=0.5]
(axis cs:0,53.6903559372885)
--(axis cs:0,48.9763107293782)
--(axis cs:1,41.3392361366099)
--(axis cs:2,36.4281202509495)
--(axis cs:3,38.4093121615589)
--(axis cs:4,42.368170770252)
--(axis cs:5,41.8134711993814)
--(axis cs:6,39.934990484003)
--(axis cs:7,39.6721603511142)
--(axis cs:8,40.082079644611)
--(axis cs:9,39.1303571998177)
--(axis cs:10,39.7785042337395)
--(axis cs:11,39.8332489485278)
--(axis cs:12,40.3914841908972)
--(axis cs:12,42.5685158091028)
--(axis cs:12,42.5685158091028)
--(axis cs:11,41.9400843848055)
--(axis cs:10,40.8348290995938)
--(axis cs:9,41.3696428001823)
--(axis cs:8,41.1645870220557)
--(axis cs:7,41.5678396488858)
--(axis cs:6,41.9783428493303)
--(axis cs:5,42.699862133952)
--(axis cs:4,44.811829229748)
--(axis cs:3,45.6306878384411)
--(axis cs:2,47.6785464157172)
--(axis cs:1,46.0674305300568)
--(axis cs:0,53.6903559372885)
--cycle;

\path [draw=mycolor3, fill=mycolor3, opacity=0.5]
(axis cs:0,54.4142135623731)
--(axis cs:0,51.5857864376269)
--(axis cs:1,48.8904526004315)
--(axis cs:2,45.0034781513626)
--(axis cs:3,40.9621800577889)
--(axis cs:4,43.1378267605562)
--(axis cs:5,43.6741483500313)
--(axis cs:6,43.6601658832979)
--(axis cs:7,43.6822897866467)
--(axis cs:8,44.5098462615449)
--(axis cs:9,44.7628940709517)
--(axis cs:10,44.5668442316016)
--(axis cs:11,44.4095468856859)
--(axis cs:12,45.014072990215)
--(axis cs:12,46.785927009785)
--(axis cs:12,46.785927009785)
--(axis cs:11,47.3771197809808)
--(axis cs:10,46.5064891017317)
--(axis cs:9,46.2304392623816)
--(axis cs:8,46.3168204051218)
--(axis cs:7,47.5377102133533)
--(axis cs:6,47.8798341167021)
--(axis cs:5,48.1858516499687)
--(axis cs:4,48.0021732394438)
--(axis cs:3,46.2911532755444)
--(axis cs:2,49.2231885153041)
--(axis cs:1,49.8228807329019)
--(axis cs:0,54.4142135623731)
--cycle;

\path [draw=mycolor4, fill=mycolor4, opacity=0.5]
(axis cs:0,46.6805435201142)
--(axis cs:0,30.6527898132191)
--(axis cs:1,29.8905965778539)
--(axis cs:2,28.6663762398116)
--(axis cs:3,34.9522462529464)
--(axis cs:4,35.2356071540155)
--(axis cs:5,34.4297676124182)
--(axis cs:6,35.9591791155706)
--(axis cs:7,35.8265040892157)
--(axis cs:8,34.5183743503458)
--(axis cs:9,33.6845516113859)
--(axis cs:10,34.1617063496655)
--(axis cs:11,33.1625970446956)
--(axis cs:12,32.3342018295038)
--(axis cs:12,36.1457981704962)
--(axis cs:12,36.1457981704962)
--(axis cs:11,36.8907362886378)
--(axis cs:10,37.5916269836679)
--(axis cs:9,38.1887817219474)
--(axis cs:8,37.2949589829875)
--(axis cs:7,36.9201625774509)
--(axis cs:6,36.9074875510961)
--(axis cs:5,37.7768990542485)
--(axis cs:4,36.2643928459845)
--(axis cs:3,41.0144204137203)
--(axis cs:2,47.1069570935217)
--(axis cs:1,46.1494034221461)
--(axis cs:0,46.6805435201142)
--cycle;

\path [draw=mycolor5, fill=mycolor5, opacity=0.5]
(axis cs:0,45.9329966244078)
--(axis cs:0,32.7336700422589)
--(axis cs:1,38.2543788256403)
--(axis cs:2,30.9394010429667)
--(axis cs:3,33.9663231013907)
--(axis cs:4,34.7492641065543)
--(axis cs:5,33.6896215861848)
--(axis cs:6,34.8025228894738)
--(axis cs:7,34.9501679055582)
--(axis cs:8,33.8350808893626)
--(axis cs:9,34.6799618326171)
--(axis cs:10,33.988420159966)
--(axis cs:11,31.9094633979643)
--(axis cs:12,31.9670459747562)
--(axis cs:12,36.1262873585771)
--(axis cs:12,36.1262873585771)
--(axis cs:11,37.3972032687024)
--(axis cs:10,37.091579840034)
--(axis cs:9,37.3000381673829)
--(axis cs:8,38.2649191106374)
--(axis cs:7,37.6631654277752)
--(axis cs:6,38.2308104438595)
--(axis cs:5,38.1237117471485)
--(axis cs:4,37.6640692267791)
--(axis cs:3,44.380343565276)
--(axis cs:2,39.8072656236999)
--(axis cs:1,45.7989545076931)
--(axis cs:0,45.9329966244078)
--cycle;

\path [draw=mycolor6, fill=mycolor6, opacity=0.5]
(axis cs:0,43.0236892706218)
--(axis cs:0,38.3096440627115)
--(axis cs:1,26.8767379792764)
--(axis cs:2,33.5893993448677)
--(axis cs:3,37.2215239378921)
--(axis cs:4,34.4025489998059)
--(axis cs:5,33.8078362594416)
--(axis cs:6,33.8411716732108)
--(axis cs:7,33.7923896931463)
--(axis cs:8,32.2689311810115)
--(axis cs:9,31.4285234954593)
--(axis cs:10,31.5369532707118)
--(axis cs:11,30.6837288029403)
--(axis cs:12,30.6169289999745)
--(axis cs:12,33.3697376666921)
--(axis cs:12,33.3697376666921)
--(axis cs:11,33.689604530393)
--(axis cs:10,34.4830467292882)
--(axis cs:9,34.5981431712073)
--(axis cs:8,34.8377354856551)
--(axis cs:7,34.4409436401871)
--(axis cs:6,34.7788283267892)
--(axis cs:5,35.4121637405584)
--(axis cs:4,36.3441176668607)
--(axis cs:3,40.4718093954413)
--(axis cs:2,39.8239339884656)
--(axis cs:1,37.7832620207236)
--(axis cs:0,43.0236892706218)
--cycle;

\addplot [mycolor1, forget plot, line width=2.5pt]
table {%
0 44.6666666666667
1 45.8766666666667
2 41.29
3 40.66
4 43.0966666666667
5 42.59
6 42.3466666666667
7 42.1066666666667
8 42.51
9 42.48
10 42.26
11 42.95
12 43.0433333333333
};
\addplot [mycolor2, forget plot, line width=2.5pt]
table {%
0 51.3333333333333
1 43.7033333333333
2 42.0533333333333
3 42.02
4 43.59
5 42.2566666666667
6 40.9566666666667
7 40.62
8 40.6233333333333
9 40.25
10 40.3066666666667
11 40.8866666666667
12 41.48
};
\addplot [mycolor3, forget plot, line width=2.5pt]
table {%
0 53
1 49.3566666666667
2 47.1133333333333
3 43.6266666666667
4 45.57
5 45.93
6 45.77
7 45.61
8 45.4133333333333
9 45.4966666666667
10 45.5366666666667
11 45.8933333333333
12 45.9
};
\addplot [mycolor4, forget plot, line width=2.5pt]
table {%
0 38.6666666666667
1 38.02
2 37.8866666666667
3 37.9833333333333
4 35.75
5 36.1033333333333
6 36.4333333333333
7 36.3733333333333
8 35.9066666666667
9 35.9366666666667
10 35.8766666666667
11 35.0266666666667
12 34.24
};
\addplot [mycolor5, forget plot, line width=2.5pt]
table {%
0 39.3333333333333
1 42.0266666666667
2 35.3733333333333
3 39.1733333333333
4 36.2066666666667
5 35.9066666666667
6 36.5166666666667
7 36.3066666666667
8 36.05
9 35.99
10 35.54
11 34.6533333333333
12 34.0466666666667
};
\addplot [mycolor6, forget plot, line width=2.5pt]
table {%
0 40.6666666666667
1 32.33
2 36.7066666666667
3 38.8466666666667
4 35.3733333333333
5 34.61
6 34.31
7 34.1166666666667
8 33.5533333333333
9 33.0133333333333
10 33.01
11 32.1866666666667
12 31.9933333333333
};
\end{axis}

\end{tikzpicture}%
    \label{fig:pwm_uniform_fill_different_speeds}}
    \hfill
    \subfloat[]{\input{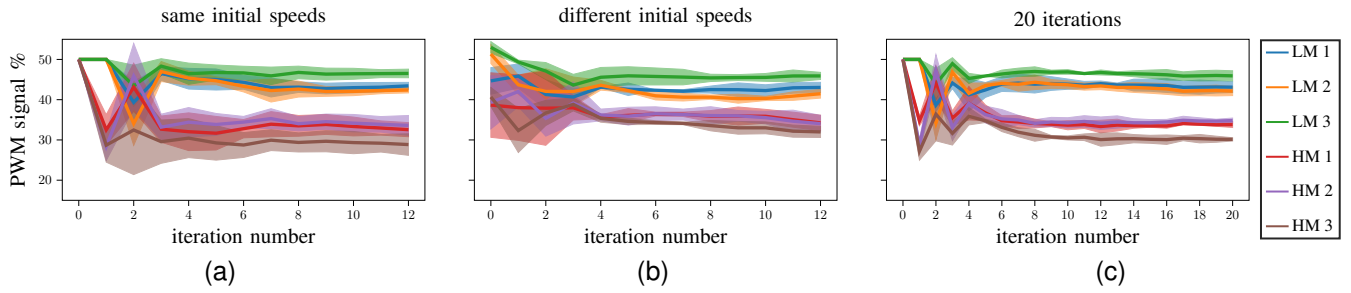}%
    \label{fig:pwm_uniform_20_fill}}

    \caption{This figure shows the PWM signals of the motors for three different tests. All three tests have the same uniform target distribution. Panel (a) shows the PWM signals in the case of starting the test with all motors running at 50\% speed and running the algorithm for twelve iterations. Panel (b) shows the average value of the PWM signals when starting the test with random initial motor speeds and running the algorithm for twelve iterations. Panel (c) demonstrates the PWM signals in the case of starting the test with all motors running at 50\% speed and running the algorithm for 20 iterations. The shaded regions indicate the standard deviation across multiple experiments (five experiments for (a), and three experiments for (b) and (c)).}%
    \label{fig:Convergence_PWM_signals}%
\end{figure*}

\subsection{Experimental results}

We showed that our algorithm is able to control the wind tunnel's motors to generate airflow matching the desired distribution. In the following, we analyze the convergence characteristics of our algorithm. All experiments presented in
Section \ref{sec:results}, started with the same initial motors' speeds, which is $50\%$  of the maximum speed, and ran for twelve iterations. In the following, we include additional experiments with more iterations and different initial conditions using only six of the seven motors with a uniform target distribution.

In the first set of tests, we assigned random initial speeds to the motors and conducted the experiment over twelve iterations. This process was repeated three times with varying initial speeds. We observed that in all three experiments, after six to seven iterations, the motors' PWM signals converged to the same values which were similar to the PWM values in the main experiment. This indicates that our algorithm successfully identified a global optimum for the PWM signals (see Figure \ref{fig:Convergence_PWM_signals} (b)).

In the second set of tests, we initialized all motors with the same initial speed ($50\%$ maximum speed) and ran the experiment for 20 iterations to check the long-term behavior of the algorithm. We note that the PWM signals converged after six iterations and remained stable (see Figure \ref{fig:Convergence_PWM_signals} (c)). 

\subsection{Proof of Prop. 1}
We define the error at iteration $k$ as $e_k = F(U_k) - F_\mathrm{des}$ and the update step as $\Delta U_k = U_{k+1} - U_k$. From Algorithm \ref{alg:online_learning_algorithm}, the update step is:
\begin{equation}
    \Delta U_{k} = -\lambda \frac {\partial \Phi_{\mathrm{pr}}}{\partial U} \biggm\vert_{U_{k}}^{+} e_{k}.
\label{eq:algorithm_update}
\end{equation}
As a result of assumption \eqref{eq:assumptionsA}, the $L$-smoothness of $\zeta(U)$ yields:
\begin{align} \label{eq:error_smooth} \begin{aligned}
\zeta(U_{k+1}) &\le \zeta(U_{k}) + \frac{\partial \zeta}{\partial U}\biggm\vert_{U_{k}} \Delta U_{k} +\frac{L}{2} \lVert \Delta U_{k} \rVert_{2}^{2}\\
&= \zeta(U_{k}) - \lambda \frac{\partial \zeta}{\partial U}\biggm\vert_{U_{k}} \frac{\partial \Phi_{\mathrm{pr}}}{\partial U}\biggm\vert_{U_{k}}^{+} e_{k} \\
& \quad +\frac{L}{2} \lambda^{2} \LNorm{\frac{\partial \Phi_{\mathrm{pr}}}{\partial U}\biggm\vert_{U_{k}}^{+} e_{k} }_{2}^{2}.
\end{aligned}
\end{align}
From the definition of $\zeta(U)$, we have:
\begin{equation}
    \frac{\partial \zeta}{\partial U}\biggm\vert_{U_{k}} = e_{k}^\top \frac{\partial F}{\partial U}\biggm\vert_{U_{k}} .
\label{eq:zeta_der}
\end{equation}
Substituting \eqref{eq:zeta_der} into \eqref{eq:error_smooth} yields:
\begin{align} \label{eq:convergance_2} \begin{aligned}
\zeta(U_{k+1}) &\le \zeta(U_{k})  +\frac{L}{2} \lambda^{2} \LNorm{\frac{\partial \Phi_{\mathrm{pr}}}{\partial U}\biggm\vert_{U_{k}}^{+} e_{k}}_{2}^{2}\\
& \quad - \lambda e_{k}^{\top}\frac{\partial F}{\partial U}\biggm\vert_{U_{k}} \frac{\partial \Phi_{\mathrm{pr}}}{\partial U}\biggm\vert_{U_{k}}^{+} e_{k}.
\end{aligned}
\end{align}
We can write this expression symmetrically with respect to $e_k$:
\begin{align} \label{eq:convergance_3} \begin{aligned}
\zeta(U_{k+1}) &\le \zeta(U_{k}) + e_{k}^\top \Biggl[\! -\frac{\lambda}{2} \!\left( \frac{\partial F}{\partial U}\frac{\partial \Phi_{\mathrm{pr}}}{\partial U}^{+}\! +\! \frac{\partial \Phi_{\mathrm{pr}}}{\partial U}^{+^\top}\!\frac{\partial F}{\partial U}^\top\! \right) \\
& \quad + \frac{L}{2}\lambda^2 \frac{\partial \Phi_{\mathrm{pr}}}{\partial U}^{+^\top}\frac{\partial \Phi_{\mathrm{pr}}}{\partial U}^{+} \Biggr] e_{k}.
\end{aligned}
\end{align}
Applying the matrix inequality assumption \eqref{eq:assumptionsB} to the bracketed term in \eqref{eq:convergance_3} yields:
\begin{align} \label{eq:convergance_4} \begin{aligned}
\zeta(U_{k+1}) &\le \zeta(U_k) - c \lambda e_{k}^\top \frac{\partial F}{\partial U}\frac{\partial F}{\partial U}^\top e_{k} \\
&= \zeta(U_k) - c \lambda \LNorm{\frac{\partial \zeta}{\partial U}}_2^2.
\end{aligned}
\end{align}
Applying assumption \eqref{eq:assumptionsC} to the gradient norm in \eqref{eq:convergance_4} yields:
\begin{equation}
    \zeta(U_{k+1}) \le \zeta(U_k) - 2\mu c \lambda (\zeta(U_k) - \zeta^*).
\end{equation}
Subtracting $\zeta^*$ from both sides of the inequality gives:
\begin{align} \label{eq:convergance_5} \begin{aligned}
\zeta(U_{k+1}) - \zeta^* &\le \zeta(U_k) - \zeta^* - 2\mu c \lambda (\zeta(U_k) - \zeta^*) \\
&= (1 - 2\mu c \lambda) (\zeta(U_k) - \zeta^*).
\end{aligned}
\end{align}
By induction, applying this relation over $k$ iterations yields the desired result:
\begin{equation}
    \zeta(U_k) - \zeta^* \le (1 - 2\mu c \lambda)^k (\zeta(U_0) - \zeta^*).
\end{equation}
This concludes the proof.


\bibliographystyle{IEEEtran}
\bibliography{literature}

\end{document}